\documentclass[runningheads]{llncs}
\usepackage[hidelinks, colorlinks=true, linkcolor=blue,citecolor = blue ,breaklinks = true]{hyperref}
\usepackage[T1]{fontenc}
\usepackage{graphicx}
\usepackage{multirow}

\usepackage{colortbl}
\usepackage{booktabs}
\usepackage{ulem}
\usepackage{pifont}
\usepackage{amsfonts,amssymb,amsmath}
\usepackage{dsfont}
\usepackage{float}
\usepackage{subcaption}
\usepackage{amsmath}
\usepackage{multirow}
\usepackage{graphicx}  
\usepackage{booktabs}  
\usepackage{amssymb}   
\usepackage{arydshln}

\usepackage{pgfplots}
\usepackage{pgfplotstable}
\usepackage{tikz}

\usepackage{caption}

\begin{document}
%
\title{Robust Multimodal Learning for Ophthalmic Disease Grading via Disentangled Representation}
\makeatletter\renewcommand{\inst}[1]{\textsuperscript{#1}}\makeatother
\author{
    Xinkun Wang\inst{1}\textsuperscript{*} \and 
    Yifang Wang\inst{1}\textsuperscript{*} \and 
    Senwei Liang\inst{1}\textsuperscript{*} \and
    Feilong Tang\inst{1,2}\textsuperscript{†} \and
    Chengzhi Liu\inst{3} \and
    Ming Hu\inst{2} \and 
    Chao Hu\inst{4} \and 
    Junjun He\inst{5} \and 
    Zongyuan Ge\inst{2}\textsuperscript{†} \and 
    Imran Razzak\inst{1}\textsuperscript{†}
}

\authorrunning{X. Wang et al.}

\institute{
    \inst{1} MBZUAI, United Arab Emirates \\
    \inst{2} Monash University, Australia \\
    \inst{3} Liverpool University, United Kingdom \\
    \inst{4} China Unicom (Shanghai) Industrial Internet Co., Ltd., China \\
    \inst{5} Shanghai AI Lab, China \\
}

\maketitle              
\renewcommand{\thefootnote}{}
\footnotetext{\textsuperscript{*}These authors contribute equally}
\footnotetext{\textsuperscript{†}Corresponding authors: \texttt{Zongyuan.Ge@monash.edu}, \texttt{Imran.razzak@mbzuai.ac.ae},
\texttt{Feilong.Tang@monash.edu}
}
\renewcommand{\thefootnote}{\arabic{footnote}}

\begin{abstract}
Ophthalmologists often rely on multimodal data to improve diagnostic precision. However, data on complete modalities are rare in real applications due to a lack of medical equipment and data privacy concerns. Traditional deep learning approaches usually solve these problems by learning representations in latent space. However, we highlight two critical limitations of these current approaches: \textit{(i)} Task-irrelevant redundant information existing in complex modalities (\textit{e.g.,} massive slices) leads to a significant amount of redundancy in latent space representations. \textit{(ii)} Overlapping multimodal representations make it challenging to extract features that are unique to each modality. To address these, we introduce the \textbf{E}ssence-Point and \textbf{D}isentangle \textbf{R}epresentation \textbf{L}earning (\textbf{EDRL}) strategy that integrates a self-distillation mechanism into an end-to-end framework to enhance feature selection and disentanglement for robust multimodal learning. Specifically, Essence-Point Representation Learning module selects discriminative features that enhance disease grading performance. Moreover, the Disentangled Representation Learning module separates multimodal data into modality-common and modality-unique representations, reducing feature entanglement and enhancing both robustness and interpretability in ophthalmic disease diagnosis. Experiments on ophthalmology multimodal datasets demonstrate that the proposed EDRL strategy outperforms the state-of-the-art methods significantly. Code is available at \href{https://github.com/xinkunwang111/Robust-Multimodal-Learning-for-Ophthalmic-Disease-Grading-via-Disentangled-Representation}{GitHub Repository}.


\keywords{Missing Modality \and Multi Modality \and Ophthalmic Disease}
\end{abstract}
%
%
\section{Introduction}
\label{sec:intro}
In recent years, using multimodal data sources has become a common method to enhance diagnostic accuracy for ophthalmic diseases \cite{watanabe2022combining,wang2023learnable,lam2024performance}. In these methods, Optical Coherence Tomography (OCT) and Retinal Fundus Imaging are typically used modalities \cite{mleppat2019directional,meleppat2021in}. Existing methods primarily focus on modality feature fusion, employing spatial and channel attention~\cite{zheng2023casf,woo2018cbam} or evidence fusion models with the inverse gamma prior distribution~\cite{ZOU2024103214}.

\begin{figure}[t]
    \centering
    \includegraphics[width=\linewidth]{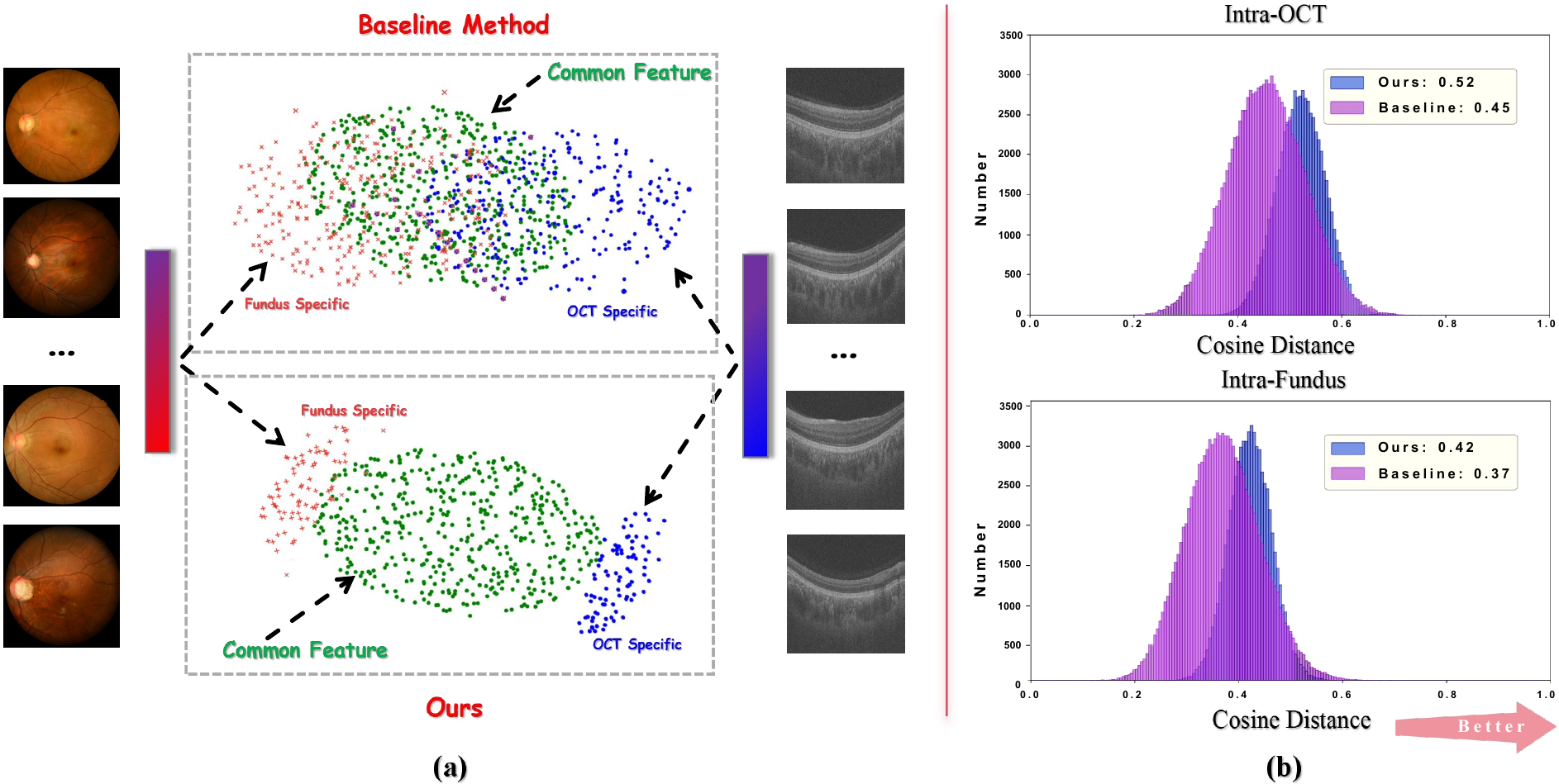}
    \caption{Overview of feature representation analysis. Baseline methods employ Vision Transformer \cite{Dosovitskiy2020} to extract and concatenate features from both modalities. (a) t-SNE \cite{VanDerMaaten2008} visualization illustrates the distribution of modality-specific and modality-common features, comparing a baseline method with our strategy. (b) Cosine distance quantifies feature separability by measuring how effectively feature from different samples are distinguished within each modality.}
    \label{fig:redundancy}
\end{figure}


Although numerous representation learning methods have been developed to address missing modality scenarios, two major issues still exist. \textbf{(1) Task-irrelevant Redundant Information}: In the absence of precise annotations, such as patch-wise labeling for regions affected by ophthalmic diseases in fundus and OCT images~\cite{huang1991optical,muller2019ophthalmic}, feature representations often contain both task-relevant and irrelevant information relevant to the task~\cite{hosseini2023computational,udandarao2023sus}. As shown in Fig.~\ref{fig:redundancy} (b), the baseline method exhibits lower cosine distance between distinct samples, indicating an insufficient ability to capture distinguishable features and leading to lower grading performance~\cite{rippel2015metric,wang2020understanding}. 
\textbf{(2) Overlapping multimodal representations:} Most methods~\cite{chen2021multimodal,xu2023multimodal,liang2022mind} that focus on cross-modality common representation extraction lead to feature representations of different modalities that have a substantial amount of cross-modal shared information. As shown in Fig.~\ref{fig:redundancy} (a), there exists a significant overlap between features of different modalities, hindering the model from utilizing the modality-unique information for diagnosis~\cite{Wang2024Decoupling,xiong2021ask}.

To this end, we propose \textbf{E}ssence-point and \textbf{D}isentangle \textbf{R}epresentation \textbf{L}earning (EDRL) framework. Specifically, the Essence-point Representation Learning (EPRL) module models the essence-points that guide the selection of discriminative information within each modality, effectively reducing task-irrelevant redundant representation. For modality feature disentanglement, the Disentangle Representation Learning (DiLR) module decomposes feature embeddings into modality-common and modality-unique components. It aligns shared information by driving the cross-correlation matrix toward a unit form while ensuring that modality-specific features remain distinct by approaching a zero form. Furthermore, we incorporate self-distillation between two pipelines (missing vs. complete modalities), using the complete pipeline to guide missing information reconstruction and enhance robustness under incomplete-modality conditions. Consequently, the EDRL framework minimizes redundancy, reduces inter-modality overlap, and enhances discriminative and generalizable multimodal learning.

Overall, our contributions are threefold. \textit{(i)} We propose EPRL framework to reduce intra-modal redundancy by modeling prototypes for discriminative instance selection and incorporating self-distillation mechanism. \textit{(ii)} We introduce DiLR to disentangle multimodal features into independent modality-unique and modality-common representation. \textit{(iii)} Experiments on three ophthalmology multimodal datasets demonstrate the effectiveness of the EDRL strategy.





\begin{figure}[t]
    \centering
    \includegraphics[width=\linewidth]{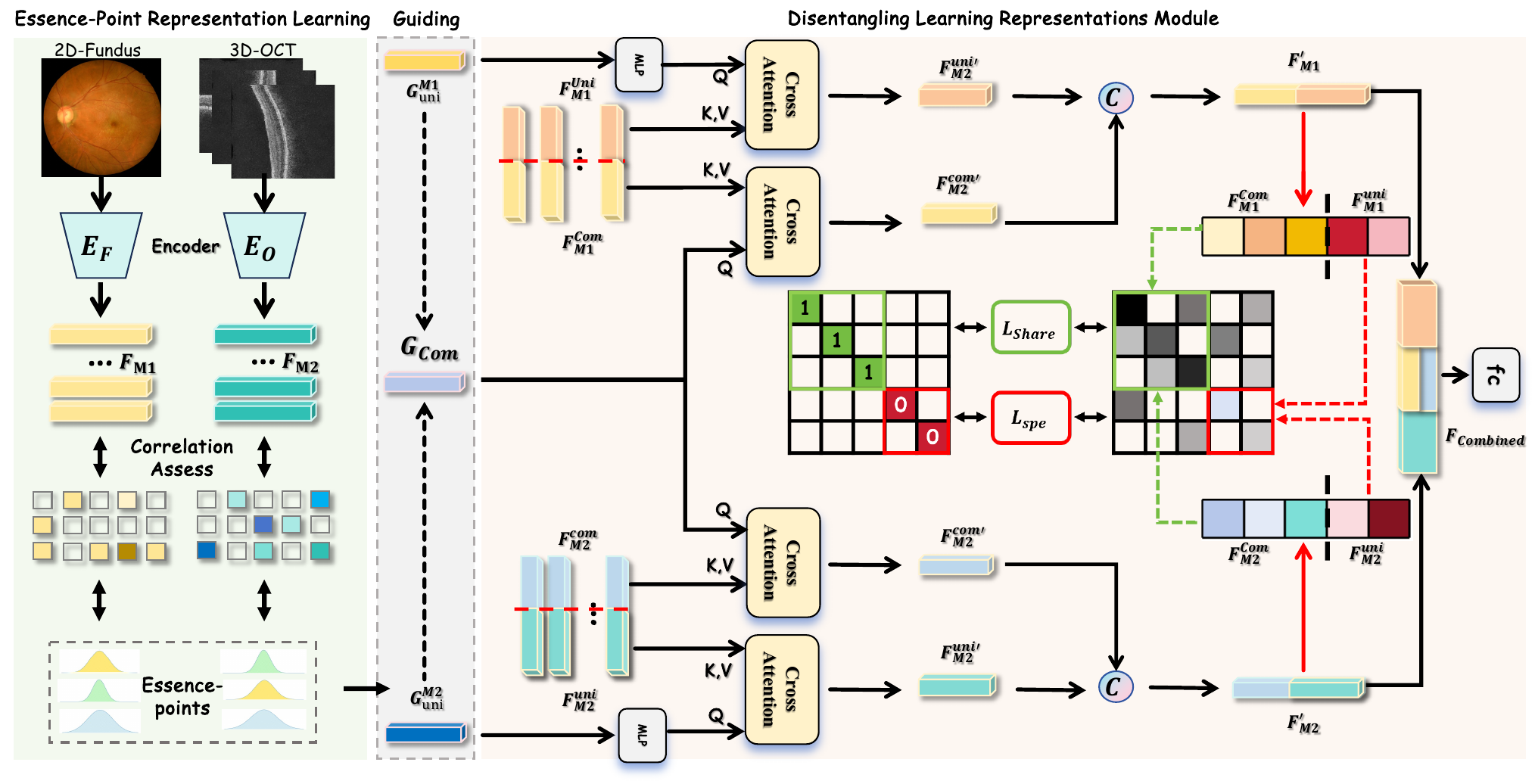}
    \caption{Illustrates our proposed \textbf{EDRL} framework, comprising two key modules: \textbf{EPRL} and \textbf{DiLR}. The EPRL module maintains series of essence-points to extract discriminative features (e.g., \(F_{M1}\) and \(F_{M2}\)) from each modality. The DiLR module disentangles these features into independent modality-common (\(F_{Com}\)) and modality-unique (\(F_{Uni}\)) representations, leveraging attention mechanisms to align shared information while preserving modality-specific characteristics. \(F_{Com}\) and \(F_{Uni}\) are then concatenated into (\(F_{Combined}\)) for grading tasks.}
    \label{fig:framework}
\end{figure}

\section{Methods}
\subsection{Problem Formulation}

We represent $ A = \{ \mathbf{a}_j, b_j \}_{j=1}^K $ as a multimodal dataset with $ K $ patient samples. Each ophthalmological sample $ \mathbf{a}_j $ consists of $ L $ inputs from different modalities, written as $ \mathbf{a}_j = \{ \mathbf{a}_j^l \}_{l=1}^L $, where $ L $ denotes the number of modalities and $ b_j \in \{ 1, 2, \dots, D \} $ is the label for $ \mathbf{a}_j $, with $ D $ being the number of grading categories.

In this work, we propose an EDRL framework to address the challenge of missing modalities, particularly for solving (1) Inter-modality missing, where one modality is completely missing, and (2) Intra-modality missing, where natural noises are added into the modalities. To reduce "Task-irrelevant Redundant Information", we propose Essence-Point Representation Learning (EPRL) module to select the task-relevant representation. Subsequently, to deal with "overlapped multimodal representations", we propose Disentangling Learning Representations (DiLR) module to generate independent modality-unique and modality-common representation. The overall framework is shown in Fig.~\ref{fig:framework}. 

\subsection{EPRL: Essence-Point Representation Learning}
We propose EPRL to filter out information in the feature map that is indiscriminative to the ohthalmic disease grading task. Since the task-discriminative information follows conditional distributions given modality type \(m\) and class label \(c\), EPRL maintains $m \times c$ learnable essence-points \( E_{m}^{c} \) for each \(m\) and \(c\), aiming to model discriminative information distribution given \(m\) and \(c\). To guide essence-point learning during training process, we need to match these essence-points with the feature representation based on \(m\) and \(c\). Such process can be implemented by the matching loss function \( L_{\text{Matching}} \). For each modality \(m\), the loss encourages the feature representation \(F_{M}^{c}\) to be aligned with their corresponding essence-points \( E_{m}^{c} \), while simultaneously minimizing their similarity with essence-points from other classes. Suppose \( N \) is the batch size and \( K \) is the total number of classes, \( L_{\text{Matching}} \) with cosine similarity is defined as:
\vspace{-0.3cm}
\begin{equation}
   L_{\text{Matching}} = - \frac{1}{N} \sum_{i=1}^{N} \left( \text{Sim}(\mathbf{F}_{M}^{c}, \mathbf{E}_{M}^{c}) - \frac{1}{2K-1} \sum_{j \neq c}^{2K-1} \text{Sim}(\mathbf{F}_{M}^{c}, \mathbf{E}_{M}^{j}) \right).
\end{equation}
During the inference, due to the lack of guidance by the label, EPRL will conduct the correlation assessment and select the highest similarity essence-point.

The \textit{\textbf{Guiding}} process aims to generate guiding tokens \( G_{uni}^{M} \) that direct the multi-modal representations \( \{ F_{M1}, F_{M2} \} \) to focus on task-relevant regions while eliminating unrelated information. Assuming that the essence-points follow a Gaussian distribution in each modality, we first employ an MLP to predict the mean and variance of the distributions for the essence-points in label \(c\), denoted as \( N_{\text{oct}}^{c} \) and \( N_{\text{fundus}}^{c} \). The guiding tokens \( G_{uni}^{M1} \) and \( G_{uni}^{M2} \) are sampled from them, respectively. Subsequently, to obtain the cross-modality shared representation, we use the Product-of-Experts \cite{hinton2002training} to generate the joint distribution \( N_{\text{Joint}}^{c} \) based on the two individual distributions \( N_{\text{oct}}^{c} \) and \( N_{\text{fundus}}^{c} \) by assuming independence. Then,  guiding token \(G_{com}\) is randomly sampled from \( N_{\text{Joint}}^{c} \). 

\subsection{DiLR: Disentangling Learning Representations Module}
To decouple the representation into independent modality-unique and modality-common features, we introduce the DiLR module. We first decompose the feature embeddings in EPRL  \( \mathbf{F}_{M1}, \mathbf{F}_{M2} \in \mathbb{R}^D \) into two distinct parts: \( \mathbf{F}_{M}^{com} \in \mathbb{R}^{D_c} \), \( \mathbf{F}_{M}^{uni} \in \mathbb{R}^{D_u} \), where \( D_c + D_u = D \). We assume \( D_c \) represents the common features across the modalities, while \( D_u \) captures the modality-specific features. Subsequently, the guiding tokens \( G_{uni}^{M1} \), \( G_{uni}^{M2} \), and \( G_{com} \) from EPRL are used to instruct the task-discriminative information selection in \( F_{M1} \) and \( F_{M2} \) through cross-attention. Its output, with task-unrelated information removed, \( \mathbf{F}_{M1}^{com'} \) and \( \mathbf{F}_{M2}^{com'} \), should remain highly similar, while \( \mathbf{F}_{M1}^{uni'} \) and \( \mathbf{F}_{M2}^{uni'} \) are expected to be decorrelated from each other.

With this in mind, we measure the similarity of two embeddings \( \mathbf{F}_{M1}, \mathbf{F}_{M2} \in \mathbb{R}^D \) through the corrleation matrix:

\vspace{-0.1cm}
\begin{equation}
    c_{ij} = \frac{\sum_b \mathbf{F}_{M1,b,i} \mathbf{F}_{M2,b,j}}{\sqrt{\sum_b (\mathbf{F}_{M1,b,i})^2} \sqrt{\sum_b (\mathbf{F}_{M2,b,j})^2}},
\end{equation}
where \( b \) indexes batch samples, and \( i, j \) indexes the dimension of the embeddings. \( \mathbf{C}_{ij} \in \mathbb{R}^{D \times D} \) is a square matrix with values ranging from -1 to 1. In \( \mathbf{C}_{ij}\), we select the submatrix \( \mathbf{C}_{com} \in \mathbb{R}^{D_c \times D_c} \) that only utilizes the common dimensions from \( \mathbf{F}_{M1} \) and \( \mathbf{F}_{M2} \) to denote the similarity between two common features \( \mathbf{F}_{M1}^{com} \) and \( \mathbf{F}_{M2}^{com} \). Since \( \mathbf{F}_{M1}^{com} \) and \( \mathbf{F}_{M2}^{com} \) should remain high in similarity, \( \mathbf{C}_{com} \) should approach the identity matrix. \( \mathbf{C}_{uni} \) is expected to approximate a target matrix with zero diagonal conversely. Thus, the common loss and unique loss are respectively defined as:

\vspace{-0.1cm}
\begin{equation}
{L}_{com} = \sum_i \left( 1 - c_{cii} \right)^2 + \lambda_c \cdot \sum_i \sum_{j \neq i} c_{cij}^2,
\end{equation}
\vspace{-0.1cm}

\vspace{-0.1cm}
\begin{equation}
    {L}_{uni} = \sum_i c_{uii}^2 + \lambda_u \cdot \sum_i \sum_{j \neq i} c_{uij}^2.
\end{equation}

To calculate these losses, we design a realignment network. \( \mathbf{F}_{M}^{Uni} \) conducts a self-attention process to extract finer-grained features. An average operation is then employed to squeeze \( \mathbf{F}_{M}^{Uni} \). For extracting the common information from both modalities, we utilize the shared features sampled from EPRL network as the guiding token (query), while \( \mathbf{F}_{M1}^{com} \) and \( \mathbf{F}_{M2}^{com} \) serve as key and value for two cross-attention modules respectively to allow the model to extract task-related common features. Subsequently, \( \mathbf{F}_{M}^{Uni} \) and \( \mathbf{F}_{M}^{Com} \) are concatenated as \( F_{M1} \) and \( F_{M2} \) for further computation of the correlation matrix and its loss.

During inference, the representations \( F_{M1} \) and \( F_{M2} \) are concatenated (with the common part \( \mathbf{F}_{M1}^{Uni} \) and  \( \mathbf{F}_{M2}^{Uni} \) directly added) to form a combined feature \( F_{Combined} \), which is passed through an MLP to complete the classification task.

\subsection{Unified Self-Distillation Mechanism}
Specifically, feature-level and logits-level consistency are employed to guide the model towards generating more accurate representations for incomplete modalities. For feature distillation, we employ Maximum Mean Discrepancy loss to minimize the discrepancy between combined features \(F_{combine}^{miss}\) and \(F_{combine}^{complete}\). 
\begin{equation}
L_{\text{features}} = \frac{1}{B} \sum_{j=1}^{b} \hat{D}_T(F_{combine}^{miss}, F_{combine}^{complete}),
\end{equation}
where 
\begin{equation}
D_T(x, y) \triangleq \|\mathbb{E}_x[\varphi(X_1)] - \mathbb{E}_y[\varphi(X_2)]\|_T^2,
\end{equation}
where \( \varphi(\cdot) \) is a feature transformation, and \(T\) is the Reproducing Kernel Hilbert Space \cite{BerlinetThomas2004,Aronszajn1950,Okutmustur2005}. 
For logits distillation, we apply Jensen-Shannon (JS) divergence \cite{li2024unified} to minimize the difference between the logits of different modality-missing cases:
\vspace{-0.1cm}
\begin{equation}
    D_{\text{JS}}(p_1 \parallel p_2) = \frac{1}{2}\left(D_{\text{KL}}(p_1 \parallel q) + D_{\text{KL}}(p_2 \parallel q)\right),
\end{equation}
where \(q\) represents the average distribution of the logits, and the corresponding logits distillation loss is:

\begin{equation}
L_{\text{logits}} = D_{\text{JS}}(MLP(F_{Combibed}^ {1}) \parallel MLP(F_{Combined}^{2})).
\end{equation}
\vspace{-0.1cm}


\section{Experiment} \label{sec:exp}
\subsection{Datasets}
We assess the proposed framework using three public multimodal datasets:  three subsets from Harvard-30k \cite{luo2024eye}, including Harvard-30k AMD, Harvard-30k DR, and Harvard-30k Glaucoma, which focus on Age-related Macular Degeneration (AMD), Diabetic Retinopathy (DR), and Glaucoma. Harvard-30k subsets provide annotations with a four-class grading system for AMD and a two-class system for glaucoma and DR, with fundus images sized at 448 × 448 and OCT images sized at 200 × 256 × 256 (200 indicating the number of OCT slices).

\vspace{-0.3cm}
\subsection{Comparison with State-of-the-art Methods}
\vspace{-0.3cm}
We evaluate the performance of our model against three state-of-the-art multi-modality fusion techniques, as detailed in Table \ref{tab:Comparison}. For the purpose of comparison, we established a baseline method by utilizing Vision Transformer \cite{Dosovitskiy2020} and UNETR \cite{hatamizadeh2022unetr} as the backbones for the Fundus and OCT modality. The feature maps from both modalities are then simply concatenated and directly used for classification. The other compared methods include: (1) B-IF, which adopts an early fusion strategy;  (2) M\(^{2}\)LC \cite{woo2018cbam}, combining channel and spatial attention for enhanced feature integration; (3)  IMDR \cite{liu2025incomplete}, a strategy based on simple mutual information loss to implement cross-modality decoupling. The comparison is conducted in three settings, i.e., (1) complete modality,(2) noisy modality (3) one modality missing.

\begin{figure}[t!]
    \centering
    \includegraphics[width=\linewidth]{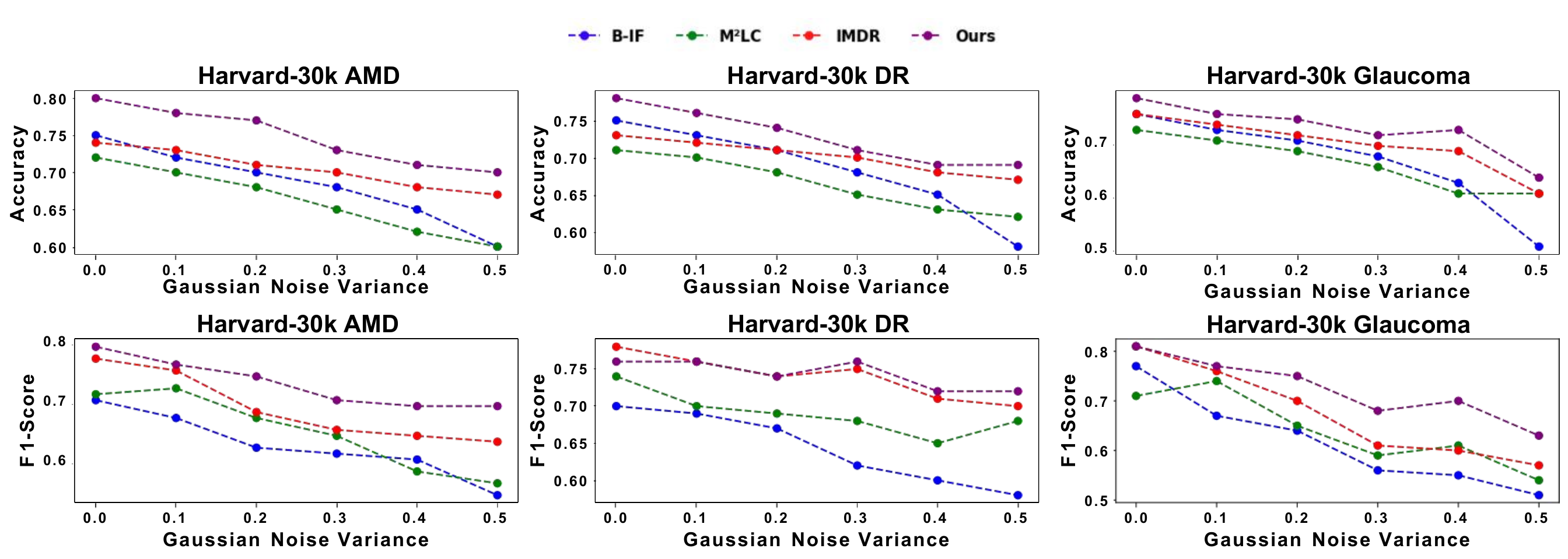}
    \vspace{-0.3cm}
    \caption{A comprehensive evaluation of performance across different missing data rates within the context of intra-modality incompleteness.}
    \label{fig:Missing rate}
\end{figure}

\noindent
\textbf{Complete Modality and noisy modality Setting.}
In the ideal scenario without any missing or noise, our model achieves the best performance among the models we test. Building upon this, we also test our approach under conditions where various Gaussian noise with different variance is introduced to each modality (In Fig \ref{fig:Missing rate}). As the noise level increases, a clear performance decline is observed in all models, emphasizing the challenges posed by data loss within a single modality on the stability of multimodal representations. Despite this, our method demonstrates exceptional robustness, particularly in scenarios with high levels of noise, consistently outperforming the other models.

\begin{table*}[t]
    \captionsetup{position=above}
    \caption{Our model is benchmarked against existing methods on the Harvard-30k dataset across three conditions: OCT missing, Fundus missing, and complete modality. The top-performing results are emphasized in bold and highlighted.}
    \label{tab:Comparison}
    \centering
    \resizebox{0.99\textwidth}{!}{%
    \begin{tabular}{c|cc|ccc|ccc|ccc}
        \toprule
        \multirow{2}{*}{\textbf{Method}} 
        & \multicolumn{2}{c|}{\textbf{Dataset}} 
        & \multicolumn{3}{c|}{\textbf{AMD}} 
        & \multicolumn{3}{c|}{\textbf{DR}} 
        & \multicolumn{3}{c}{\textbf{Glaucoma}} \\
        \cmidrule(r){2-12}
        & \multicolumn{2}{c|}{\textbf{Modality}} 
        & \textbf{OCT} & \textbf{Fundus} & \textbf{Both} 
        & \textbf{OCT} & \textbf{Fundus} & \textbf{Both}
        & \textbf{OCT} & \textbf{Fundus} & \textbf{Both}  \\ \midrule \hline
 & \multicolumn{2}{c|}{\textbf{ACC}} & 65.07 & 72.92 & 70.87 & 70.53 & 73.81 & 74.07 & 65.69 & 73.02 & 73.35 \\
Baseline & \multicolumn{2}{c|}{\textbf{AUC}} & 69.88 & 75.38 & 81.06 & 69.94 & 79.11 & 78.73 & 69.86 & 75.35 & 74.53 \\
& \multicolumn{2}{c|}{\textbf{F1}} & 69.64 & 72.28 & 70.83 & 62.01 & 70.46 & 71.17 & 70.91 & 72.31 & 71.64 \\ \midrule

& \multicolumn{2}{c|}{\textbf{ACC}} & 69.57 & 72.35 & 73.17 & 69.05 & 73.62 & 76.36 & 69.64 & 73.39 & 73.39 \\
B-IF & \multicolumn{2}{c|}{\textbf{AUC}} & 70.14 & 71.98 & 83.82 & 65.25 & 67.50 & 77.95 & 68.95 & 76.61 & 73.32 \\
& \multicolumn{2}{c|}{\textbf{F1}} & 67.45 & 70.03 & 71.25 & 67.93 & 69.68 & 75.61 & 67.18 & 72.47 & 72.11 \\ \midrule

& \multicolumn{2}{c|}{\textbf{ACC}} & 68.97 & 73.24 & 74.93 & 67.20 & 73.04 & 75.21 & 67.70 & 72.78 & 74.98 \\ 
M²LC & \multicolumn{2}{c|}{\textbf{AUC}} & 72.23 & 72.67 & 82.39 & 65.05 & 67.89 & 79.68 & 71.22 & 70.23 & 76.45 \\ 
& \multicolumn{2}{c|}{\textbf{F1}} & 65.06 & 73.80 & 71.20 & 64.33 & 74.59 & 74.39 & 65.60 & 71.11 & 74.23 \\ \midrule

& \multicolumn{2}{c|}{\textbf{ACC}} & 70.62 & 75.17 & 79.50 & 72.62 & 76.19 & 78.57 & 71.16 & 75.54 & 77.31 \\
IMDR & \multicolumn{2}{c|}{\textbf{AUC}} & 72.69 & 80.48 & 85.09 & 74.69 & 79.07 & 85.00 & 75.07 & 78.47 & 78.98 \\
& \multicolumn{2}{c|}{\textbf{F1}} & 71.90 & 76.59 & 72.52 & 72.90 & 72.18 & 77.04 & 70.37 & 75.12 & 78.90 \\ \midrule

\rowcolor{green!10} & \multicolumn{2}{c|}{\textbf{ACC}} & \textbf{71.79} & \textbf{76.69} & \textbf{81.42} & \textbf{74.38} & \textbf{77.50} & \textbf{79.50} & \textbf{72.53} & \textbf{76.28} & \textbf{78.55} \\
\rowcolor{green!10} \textbf{Ours} & \multicolumn{2}{c|}{\textbf{AUC}} & \textbf{74.84} & \textbf{81.55} & \textbf{85.82} & \textbf{76.88} & \textbf{80.60} & \textbf{86.71} & \textbf{76.28} & \textbf{79.59} & \textbf{79.32} \\
\rowcolor{green!10} & \multicolumn{2}{c|}{\textbf{F1}} & \textbf{72.94} & \textbf{76.79} & \textbf{78.93} & \textbf{74.28} & \textbf{76.71} & \textbf{79.81} & \textbf{72.54} & \textbf{76.62} & \textbf{80.54} \\ 
        \bottomrule
    \end{tabular}}
\end{table*}

\noindent
\textbf{Inter-Modality completely missing.}
We evaluate our strategy by comparing its performance with that of the other methods under OCT missing or Fundus missing situations. Even a performance decline is observed across all models when a modality is missing, our strategy demonstrates greater robustness. Result proves our strategy has robust ability to separate multimodal features and reconstruct the missing information to serve for the grading task.

\begin{figure}[t]
\centering
\begin{minipage}{0.48\textwidth}
    \vspace{-4em}
    \centering
    \begin{table}[H]
    \captionsetup{position=above}
    \caption{Baseline: Using transformer backbone to extract two modality data and simply concatenates their features. EPRL: Our Essence-point Representation Learning. DiLR: Our Disentangling Learning Representations.}
    \vspace{1em}
    \resizebox{\textwidth}{!}{
    \begin{tabular}{c|ccc|ccc}
    \toprule
    \textbf{Variants} & Baseline & EPRL & DiLR & \textbf{ACC} & \textbf{AUC} & \textbf{F1} \\ 
    \midrule
    I   & \textbf{\checkmark}  & &  & 59.51 & 63.42 & 53.47 \\ 
    II    & \textbf{\checkmark} & \textbf{\checkmark} &  &66.58 & 70.95& 66.19 \\ 
    III    & \textbf{\checkmark} &  & \textbf{\checkmark} & 64.67 & 66.45 & 64.88 \\ 
    IV    & \textbf{\checkmark} & \textbf{\checkmark} & \textbf{\checkmark} & \textbf{69.37} & \textbf{66.39} & \textbf{57.94} \\ 
    \bottomrule
    \end{tabular}
    }
    \label{tab:ablation}
    \vspace{-1.5em}
    \end{table}
\end{minipage}
\hfill
\begin{minipage}{0.48\textwidth}
    \vspace{-4em}
    \centering
    \begin{table}[H]
    \captionsetup{position=above}
    \caption{Implementation of a comprehensive hyperparameter sensitivity analysis within the full-modality framework of the Harvard-30k dataset. Percentage \textit{(p)}: the ratio of common dimensions to total dimensionality.}
    \vspace{1em}
    \resizebox{\textwidth}{!}{
    \begin{tabular}{cccccc}
    \toprule
    \textbf{\textit{(p)}} & 0.3 & 0.4 & 0.5 &0.6 & 0.7 \\
    \midrule
    \textbf{AMD} & 79.56 & 81.42 & 79.05 & 80.23 & 73.47 \\
    \textbf{DR} & 76.87 & 78.13 & 78.75 & 79.50 & 77.50 \\
    \textbf{Glaucoma} & 77.32 & 78.55 & 77.44 & 76.35 & 77.50 \\
    \bottomrule
    \end{tabular}
    }
    \label{tab:Senstive}
    \vspace{-1.5em}
    \end{table}
\end{minipage}
\end{figure}

\begin{figure}[ht!]
    \centering
    \includegraphics[width=\linewidth]{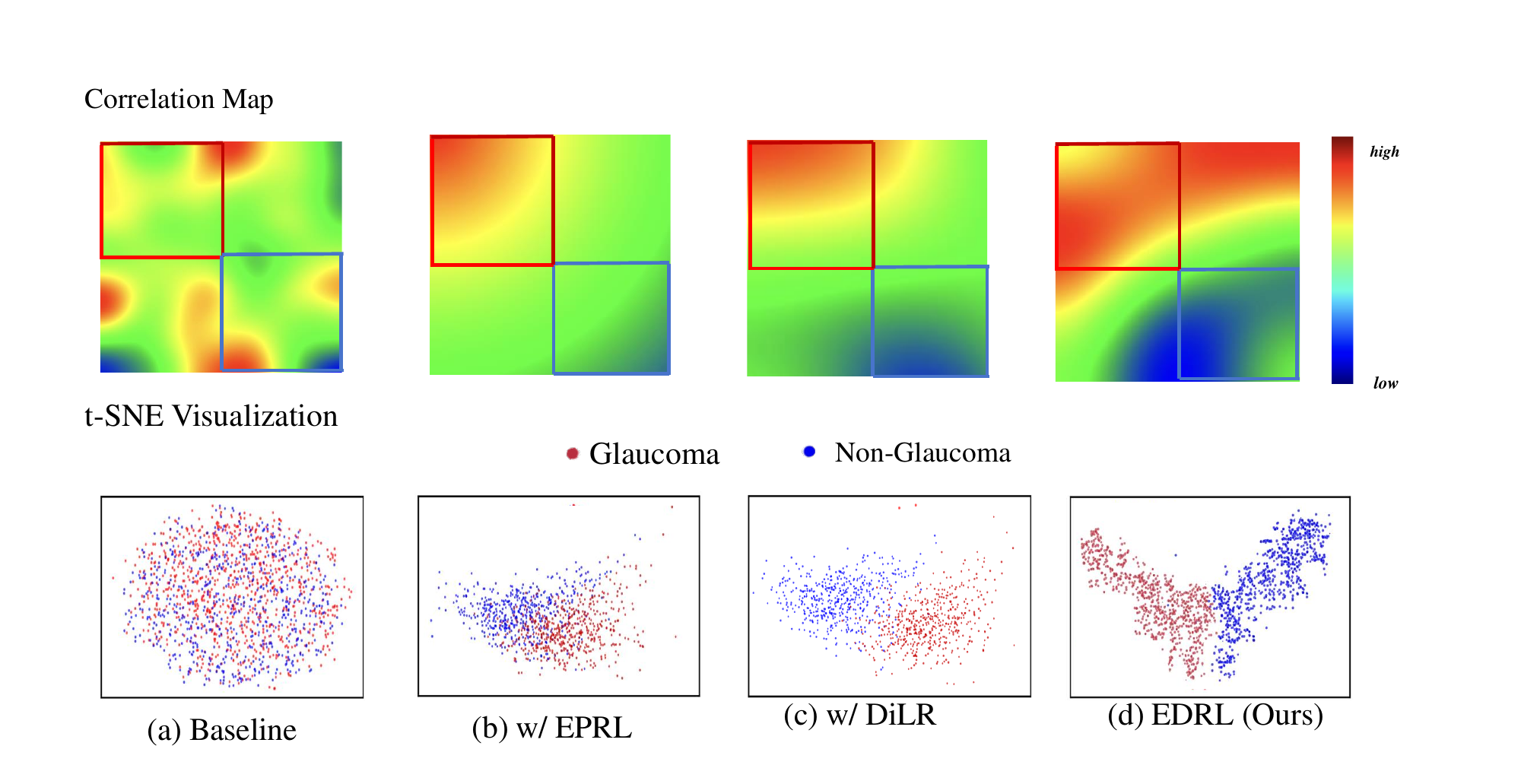}
    \vspace{-0.5cm}
    \caption{ Corrleation map and t-SNE on the Harvard-30k Glaucoma dataset. In the ideal scenario, the top-left region of the heatmap should exhibit predominantly red areas, indicating a high correlation between \(F_{M1}^{com}\) and \(F_{M2}^{com}\), while the bottom-right region should show more blue areas, signifying lower correlation between \(F_{M1}^{uni}\) and \(F_{M2}^{uni}\) }. 
    \label{fig:attention_map}
    \vspace{-1.5em}
\end{figure}


\vspace{-0.4cm}
\subsection{Ablation Study}
\vspace{-0.2cm}
\textbf{Effectiveness of each component.} To evaluate the effectiveness of our novel EPRL and DiLR, we conducted an ablation study on the test set of Harvard-30k under gaussian noise with a variance of 0.5, as shown in Table \ref{tab:ablation}. From Variant I to Variant II, we observe that the introduction of the EPRL unit effectively reduces the task-unrelated information within the modality, resulting in a significant increase in accuracy (6.5\%). From Variant I to Variant III, we find that incorporating the DiLR unit effectively decouples modality-specific and modality-unique features, leading to an accuracy increase of 5\%. Variant IV, combining both EPRL and DiLR, consistently outperforms Variants II and III, demonstrating their synergistic effect in producing highly decoupled, low-redundancy representations.


\noindent\textbf{Qualitative Results.} 
As shown in Fig.~\ref{fig:attention_map}, we visualize corrleation maps and t-SNE for four variants to assess feature decoupling and clustering. The corrleation map measures the decoupling ability of our DiLR. The baseline model  (Fig.~\ref{fig:attention_map} (a)) shows insufficient disentanglement and poorly separated clusters. Integrating EPRL (Fig.~\ref{fig:attention_map} (b)) enhances feature selection and improving clusters quality, while DiLR (Fig.~\ref{fig:attention_map} (c)) further refines modality-unique and modality-common information disentanglement and leads to better cluster separation. Finally, our EDRL (Fig.~\ref{fig:attention_map} (d)) achieves modality information disentanglement and distinct clusters, demonstrating its effectiveness in learning discriminative, modality-aware representations for grading tasks.

\noindent\textbf{Hyperparameter Sensitivity Analysis.}
To validate the robustness of our model, we conduct a series of hyperparameter sensitivity analysis in Table \ref{tab:Senstive}. In DiLR, the common dimension percentage affects performance: increasing it initially improves results, but excessive sharing impairs modality-specific information expression, causing performance decline.

\section{Conclusion}
\vspace{-0.3cm}
In multimodal ophthalmology diagnosis, two main challenges are intra-modal redundancy due to task-unrelated information and cross-modal entanglement in the latent space. To tackle these, we propose the EPRL framework to reduce redundancy, followed by the DiLR module for disentangling cross-modal features. Extensive experiments on multimodal ophthalmic datasets show that our method outperforms state-of-the-art approaches, improving interpretability.


\bibliographystyle{splncs04}
\bibliography{main}

\end{document}